\pdfoutput=1

\documentclass[11pt]{article}
\usepackage{graphicx}
\usepackage{multirow}
\usepackage[dvipsnames]{xcolor}
\usepackage{color}
\usepackage{booktabs}
\usepackage{amsmath}
\usepackage{fixltx2e}
\usepackage{acl}
\usepackage{times}
\usepackage{latexsym}
\usepackage{amssymb}
\usepackage[T1]{fontenc}

\usepackage[utf8]{inputenc}

\usepackage{microtype}
\usepackage{todonotes}

\title{Attend to the Right Context: A Plug-and-Play Module for Content-Controllable Summarization }

\author{\textbf{Wen Xiao}\footnotemark[2]\hspace{.3em}\thanks{~~Work done during an internship at Microsoft.}\qquad Lesly Miculicich\footnotemark[3]\hspace{1.9em} 
\textbf{Yang Liu}\footnotemark[3]\hspace{1.9em}  \\\textbf{Pengcheng He}\footnotemark[3]\hspace{1.9em}\textbf{Giuseppe Carenini}\footnotemark[2]\hspace{1.9em}\vspace{4pt}\\
\footnotemark[2]\hspace{.4em}University of British Columbia, Vancouver, Canada 
\\ \footnotemark[3]\hspace{.4em} Microsoft Azure AI\\
\texttt{\small{\{xiaowen3,carenini\}@cs.ubc.ca}},\\
\texttt{\small{\{leslym,yaliu10,penhe\}@microsoft.com
}}}
\begin{document}
\maketitle
\begin{abstract}

Content-Controllable Summarization generates  summaries focused on the given controlling signals. Due to the lack of large-scale training corpora for the task, we propose a plug-and-play module RelAttn to adapt any general summarizers to the content-controllable summarization task. RelAttn first identifies the relevant content in the source documents, and then makes the model attend to the right context by directly steering the attention weight.  We further apply an unsupervised online adaptive parameter searching algorithm to determine the degree of control in the zero-shot setting, while such parameters are learned in the few-shot setting. By applying the module to three backbone summarization models, experiments show that our method effectively improves all the summarizers, and outperforms the prefix-based method and a widely used plug-and-play model in both zero- and few-shot settings. Tellingly, more benefit is observed in the scenarios when
more control is needed. \footnote{Our code is available at \url{https://github.com/Wendy-Xiao/relattn_controllable_summ}}

\end{abstract}

\section{Introduction}

Abstractive summarization is the task of generating a short text containing the most relevant points for the given document. Currently, it is a widely explored task and it takes benefits from %
large-scale generative language models trained on large corpus. The state-of-the-art abstractive summarizers~\cite{pegasus,bart,primera,he2022z} achieve  good performances regarding both saliency and fluency. Following a long tradition in pre-neural query-based summarization~\cite{duc2005,rosner-camilleri-2008-multisum}, researchers have just begun to investigate the neural controllable summarizers, which are expected to generate summaries that fulfill certain constraints, either on the format, e.g. the length~\cite{lengthcontrol} and style~\cite{simplicity}, or on the content~\cite{gsum,ctrlsum}, regarding specific entities, topics, and aspects.  In this work, we focus on generating content-controllable summaries for the given documents. 

The content-controllable summarization task is essential in many practical settings. People may care about different aspects or topics even for the same news article, depending on their specific information needs, so a general summarizer is not flexible enough for such customized scenarios to generate different summaries with different requirements. Unfortunately, while there are several large naturally annotated datasets for the task of generic summarization (like news with the highlights and scientific papers with the abstracts), there are no similar large corpora for content-controllable summarization. Furthermore, annotating such a large corpus is not affordable, which calls for the zero-shot or few-shot methods. To take advantage of the general summarizers pre-trained on large-scale datasets, we propose a simple yet effective trainable plug-and-play module, RelAttn, which can effectively adapt the general summarization models to the controllable summarization task in both zero-shot and few-shot settings by steering attention weights.

Our proposal is inspired by findings in cognitive science, namely that when performing task-oriented reading comprehension tasks, like question answering or summarization, humans use \textit{selective attention} ~\cite{psycho,neuroscience,dayan2000learning,lavie2004load}, i.e., they  focus on task-relevant information while suppressing any distractions. And  intriguingly \textit{selective attention} can be further trained to improve the reader's reasoning ability\cite{psycho}.

By design, the Cross Attention in the encoder-decoder transformer models plays a similar role as \textit{selective attention}. %
For general summarizers, the Cross Attention is trained to focus on the salient content, as the task-relevant information. However, it is supposed to be the `most relevant' content %
to the given controlling signals for the content-controllable summarization task. Thus to make the model focus on the updated task-relevant information, we propose to inject an adaptable Relevance Attention component into Cross Attention.
We combine the Relevance Attention and the Cross Attention with an adaptable weight, determining how much controlling is needed. We further propose an online adaptive hyper-parameter search algorithm (OS), which can be used to determine the degree of control for each single data example in fully unsupervised settings.

One close task explored recently is \textit{guided summarization}~\cite{gsum,ctrlsum,frost}, with predicted or oracle keywords as the guidance. However, such guidance enhances the model to identify the salient content, therefore, improving the faithfulness of generated summaries, instead of making the model focus on the relevant content~\cite{entsum}.

To better align with the task, we evaluate our method on two new annotated datasets~\cite{entsum, newts}, consisting of different human-written summaries associated with different controlling signals for the same source article, which requires the model first to identify the relevant context in the source document and then generate the summaries. The experiment results show that our module improves the performances of three summarizers~\cite{bart,pegasus,ctrlsum} on both datasets, and is better than or comparable with previous prefix-based methods and a plug-and-play method on both zero-shot and few-shot settings. And more tellingly,  more benefit is observed in the scenarios when more control is needed.

\section{Related Work}
\subsection{Controlled Generation}
Recent works on Controlled Text Generation are either adding a prefix/prompt at the beginning of the input~\cite{ctrl} or adding a plug-and-play component to the large model~\cite{pplm,gedi,fudge}. 

Specifically, \citet{ctrl} introduces a large-scale conditional language model, pre-trained with control codes as a prefix derived from the structure that naturally co-occurs with raw text, e.g. the url of each document, which may specify the domain, subdomain, entities, entity relations, and even date of the document. 

In another line of work, \citet{pplm} propose the Plug-and-Play Language Model (PPLM), which combines additional attribute models with a pre-trained unconditional language model, and runs gradient ascent on the LM's hidden activations to guide the generation of the next token to satisfy the control while maintaining fluency. The attribute models can either be a simple classifier (e.g. a sentiment classifier) or even without additional parameters (e.g. computing the loss between the predictions and Bag-Of-Word of the controlling signals). Then the Plug-and-Play approach is further  extended by \citet{gedi}, who employs a generative discriminator to improve the efficiency. Even more recently, \citet{fudge} instead employs the attribute models to re-weight the output distribution of the LM considering an estimation on the controlling satisfaction of the partially generated text at each decoding step.

In this work, we compare our method with PPLM~\cite{pplm} on the datasets for the content controllable summarization task.

\subsection{Controllable Summarization}
Although the general summarization task has received much-renewed attention with the availability of powerful deep learning solutions, %
neural controlled summarization is still largely unexplored. %
Due to the nature of the task, there are mainly two kinds of controlling aspects: either about the format of generated summaries (e.g. their length~\cite{lengthcontrol} or their simplicity~\cite{simplicity} ) or about the content of the generated summaries (e.g. regarding specific entities and topics they should focus on, queries or questions they should answer).

Closely relevant to the content-controllable summarization, most recent works focus on the \textit{guided-summarization task}. Specifically, one extracts the oracle entities, keyphrases, or sentences from the ground-truth general summaries, which are then used as the guide to generate the summaries. For example, \citet{gsum} uses an additional encoder to encode the guidance signals, with partial parameters shared with the original document encoder. In another line of research, \citet{ctrlsum} propose a pre-training strategy prepending the source documents with the oracle keywords as a prefix, and similarly, \citet{frost}  train the model to first predict an entity chain before predicting the final summary. However, different from the content-controllable summarization, the injection of those keywords enhances the model to find the most important content better which improves the models' faithfulness, rather than making the model focus on relevant content in the context~\cite{entsum}.

Recently, \citet{newts} have collected a human-annotated topic-focused summarization dataset \textit{NEWTS}, sampled from the CNNDM dataset, using the topic information as the controlling signal. Similarly, \citet{entsum} have introduced a human-annotated dataset \textit{EntSUM} for the controlled summarization task with an entity as the controlling signal for each data example. Therefore, we can evaluate our proposed method on these two datasets with topic information and entities as the controlling signals.

\begin{figure*}[t]
    \centering
    \includegraphics[width=0.9\linewidth]{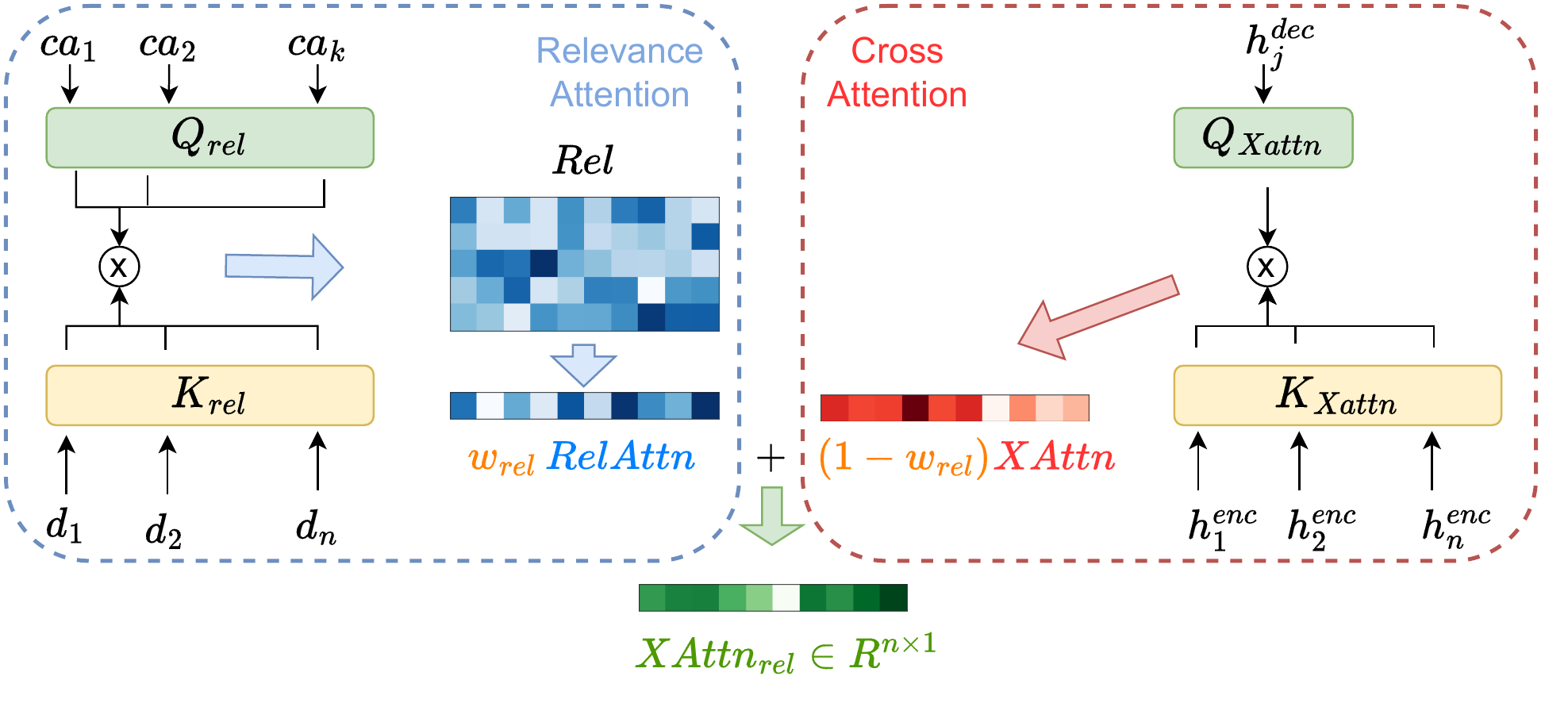}
    \caption{The illustration of the RelAttn module. The Relevance Attention is injected into the Cross Attentions with a weight $w_{rel}$, indicating how much control is conducted. In the zero-shot settings, $Q_{rel}$ and $K_{rel}$ are identical layers, while they are trainable linear layers in the few-shot scenario.}
    \label{fig:model_struct}
\end{figure*}
\section{RelAttn: A Plug-and-Play Module}
Generally, the content-controllable summarization task can be defined as follows: given a document with $n$ tokens, i.e. $D=\{d_1,d_2,...d_n\}$, and controlling aspects $CA=\{ca_1,ca_2,...ca_k\}$ with $k$ tokens, which could be a sequence of words, entities or key-phrases, indicating the content of interests, the model is supposed to generate a summary $S=\{s_1,s_2,...,s_m\}$ with $m$ tokens, regarding the given controlling aspects.

To take benefit from the large-scale pre-trained summarization models, we propose a plug-and-play module, RelAttn, adapting the general summarizers to the content-controlled summarization task. In the encoder-decoder transformer-based summarization models, the context of the source document is exclusively from the Cross Attention, which is an attention mechanism over all the input tokens in the encoder at each decoding step. Thus to guide the model to `pay more attention' to content relevant to the controlling aspects, we directly inject into the Cross Attentions a novel Relevance Attention mechanism, which explicitly represents how relevant the tokens in the source documents are to the given controlling aspects. %
Advantageously, our new RelAttn component %
can be plugged into any transformer-based sequence-to-sequence model.

In this section, we first introduce the Cross Attention in the general transformer model in Section~\ref{sec:cross_attn}, then present Relevance Attention, as well as how it is integrated into Cross Attention in Section~\ref{sec:relevance_attn}, and finally introduce how the degree of control, as a hyper-parameters, is set in both of the zero-shot and few-shot settings in Section~\ref{sec:wrel}.
\subsection{Cross Attention}
\label{sec:cross_attn}
Multi-head Attention is the key component in the transformer model~\cite{transformer}, and there are three kinds of attention in a sequence-to-sequence transformer model - Encoder Attention, Decoder Attention, and Cross Attention. 
The Cross Attention, in particular, works as a 'reference', which is used by the decoder to refer to the input sequence. As shown in Figure~\ref{fig:model_struct}(right), in each cross attention head, suppose $h_j^{dec}\in \mathbb{R}^{1\times dim}$ is the output of the previous layer of the decoder at the decoding step $j$, and $h^{enc}\in \mathbb{R}^{n\times dim}$ is the output of the last layer in the encoder, then the cross attention vector is computed as:
$$XAttn_j = Softmax(Q(h_j^{dec})\cdot K(h^{enc})^T)$$
 where $Q(\cdot)$ and $K(\cdot)$ are the query layer and key layer respectively, and $XAttn\in \mathbb{R}^{1\times n}$ indicates the attention over the input tokens, with higher attention values on input tokens that will influence more the generation of the current token.
\subsection{Relevance Attention}
\label{sec:relevance_attn}
In the general summarization models, the cross-attentions learn to identify and take more considerations of the \textit{salient content}. To adapt the models to the controllable summarization task, we propose to inject Relevance Attention into the original cross attention, which focuses on the content relevant to the given controlling aspects in the source document.

Specifically, as shown in Figure~\ref{fig:model_struct} (left), we denote the representation of each controlling aspect token $ca_j$ as $r_{ca_j}$, and the representation of each input token $d_i$ as $r_{d_i}$, then the controlling aspect with $k$ tokens and source input with $n$ tokens can be represented as $r_{ca}\in \mathbb{R}^{k\times dim}$ and $r_d\in \mathbb{R}^{n\times dim}$, respectively.\footnote{The representation can either be the embeddings of the token or the hidden states from the encoder. In this work, we simply use the embedding of each token.} The relevance attention $RelAttn$ is computed as:
\begin{eqnarray*}
Rel&=&Q_{rel}(r_{ca})\cdot K_{rel}(r_d)^T\\
RelAttn &=& Softmax(\sum_k Rel)
\end{eqnarray*}
, where $Q_{rel}(\cdot)$ and $K_{rel}(\cdot)$ are the query and key layers in the proposed RelAttn component. The resulting $Rel\in \mathbb{R}^{k\times n}$ represents the relevance of every source document token with every controlling aspect token, thus $\sum_1^k Rel \in \mathbb{R}^{1\times n}$ represents the overall relevance of the source document tokens with the given controlling aspects. In summary, $RelAttn_i$ measures how relevant the $i$-th token in the source document is with the controlling aspects, and the model is encouraged to focus more on the more relevant content to the given controlling aspects.

Note that the new parameters introduced in Relevance Attention ($Q_{rel}(\cdot)$ and $K_{rel}(\cdot)$) can be trained in few-shot and fully supervised settings. In contrast, in zero-shot settings, we use the dot product between the two representation vectors to derive the relevance between the source document and the controlling aspects, i.e. the weights in both linear layers are identical matrix, and the biases are all 0.

As shown in Figure~\ref{fig:model_struct} (middle bottom), the relevance attention matrix is then combined with the original cross attention matrices with a  relevance weight $w_{rel}$, representing how much controlling is conducted, i.e. 
\begin{eqnarray*}
XAttn_{rel}&=&w_{rel} RelAttn \\
& &+ (1-w_{rel}) XAttn\\
output&=&XAttn_{rel}\cdot V(h^{enc})
\end{eqnarray*}
, where $V(\cdot)$ represents the value layer of the encoder in the cross attention, and $XAttn\in \mathbb{R}^{1\times n}$ is the original cross attention over all the source input tokens. Thus the final output is computed using the updated cross attention $XAttn_{rel}$ with the proposed Relevance Attention included.
\begin{table*}[t!]
    \centering
    \small
    \begin{tabular}{l|cccccc}
    \toprule
     Dataset    &  \# Examples &\# Src Doc&\#Summ/Doc & Len(doc) & Len(summ) & Ctrl Asp.\\
     \midrule
     NEWTS& 4800/1200 &2388/574* &2&539&67&Topic Info\\
     EntSUM     & 734/1994&164/481&1-18&861&95&Entity\\
    \bottomrule
    \end{tabular}
    \caption{The statistics of the datasets. We randomly split EntSUM into train/test sets to evaluate the models with few-shot settings. *There are some duplicates in the NEWTS dataset, so the number of unique documents is slightly different from the expected number.}
    \label{tab:dataset_stats}
\end{table*}
\paragraph{Gaussian Smoothing:} The purpose of Relevance Attention is to find the relevant content from the documents and make the model attend to it, however, the attention vector might be excessively peaky on certain highly related words (e.g. identical words), especially in the zero-shot settings. To make the model better aware of the context of those words,  we adopt Gaussian smoothing %
in the Relevance Attention, so that attention is more spread around the relevant words. The idea is in line with the recent explorations on weight smoothing of the attentions in the transformer~\cite{relaxed_attention,smoothing_speech}.
\subsection{Determine the Degree of Control}
\label{sec:wrel}
In essence, $w_{rel}$ is a hyperparameter indicating how much influence the controlling aspects are supposed to have on the model. In this section, we %
introduce practical ways to determine the value of $w_{rel}$ in both zero-shot and few-shot settings.
\subsubsection{Zero-shot: Online Adaptive Parameter Selection}
\label{sec:os_algorithm}
Based on the property of the RelAttn module, a larger value for $w_{rel}$ %
implies that  more controlling is applied %
on the model, therefore the generated summaries are more relevant to the controlling aspects. However, similar to previous plug-and-play methods~\cite{pplm}, if the model is excessively influenced, %
it will generate repeated words or phrases, rather than a coherent summary. In previous works, hyper-parameters like $w_{rel}$ are  usually set according to the performance on a small validation set~\cite{zheng-lapata-2019-sentence}, but this is not feasible for zero-shot. %
Also, %
the attention distributions may vary for different inputs, and different degrees of control may be needed as well, which requires different $w_{rel}$ for each data example. 
We, therefore, propose an online adaptive parameter selection algorithm ($OS$) to select $w_{rel}$ for each single data example, inspired by the recent works on Minimum Bayes Risk Decoding (MBRD,~\citet{mbrd}).

The key idea is to find the `central' summary over a candidate set containing the summaries generated by the model with different $w_{rel}$.\footnote{We add some heuristic constraints to ensure the summaries in the candidate set are of reasonable quality. The details can be found in Section~\ref{sec:settings_and_implementation} }
 Specifically, for each data example $i$, we generate $n$ candidate summaries with different $w_{rel}$ to form a candidate pool $C^i_{\{w_{rel}\}}=\{S^i_{w_{rel}}\}$. And then for each candidate, we compute an alignment score with all the other candidates. We finally pick the $w_{rel}$ with the summary having the highest alignment score with other candidates, i.e.
$$\Bar{w}^i_{rel}=\mathop{argmax}_{w_{rel}} \frac{1}{|C^i_{\{w_{rel}\}}|}
\sum_{y\in \{w_{rel}\}} a(S^i_y,S^i_{w_{rel}})$$. In this way, the chosen value for $w_{rel}$ is expected to have a more balanced influence on the model for the target data example.

\subsubsection{Few-shot}
In the few-shot setting, we have even more flexibility to further make the model learn to set different $w_{rel}$ for each head. Specifically, in each head, we apply a linear layer mapping the concatenation of the representations of source documents and controlling signals to a real number, followed by a sigmoid activation function to convert the number to probability space.

\section{Datasets and Baselines}

 In this section, we will introduce the datasets and backbone models we use in the experiments.
\subsection{Datasets}
We use two datasets for the experiments, controlling the content of generated summaries by topic information and entities. The detailed statistics of the datasets can be found in Table.\ref{tab:dataset_stats}.

\paragraph{NEWTS~\cite{newts}} is a topic-oriented summarization dataset, for each article, there are two human-written summaries regarding two different topics, represented by a sequence of topic words and phrases, and a topic sentence. When building the dataset, they first obtain the topics using the LDA model~\cite{lda} on the CNNDM dataset~\cite{cnndm}, and  select the articles with two strong coherent topics, then the human annotators are asked to write two summaries regarding both topics (given by the top frequency words and manually written phrases) for each article. Based on how the dataset is built,
the topic words and phrases do not necessarily appear in both the source article and human-written summaries. For simplicity, we only use the topic words in our experiments.

\paragraph{EntSUM~\cite{entsum}} is an entity-centric summarization dataset, where each data example contains a document, an entity extracted from the document, and a human-written summary regarding the given entity. The given entity is not necessarily %
central in the source document. As the original dataset is test-only, for the purpose of hyper-parameter search and few-shot experiments, we randomly split the original dataset into train/test sets.

We show examples from both datasets in Appendix~\ref{sec:dataset_details_appendix}.

\subsection{Models}
To evaluate the effectiveness of our method, we use three pre-trained models as the backbone model, including a pre-trained model for general summarization (PEGASUS), a model fine-tuned for the general summarization task (BART-CNNDM, or in short BART\footnote{All `BART' mentioned in this paper refer to `BART-CNNDM'}), and a model trained for conditional summarization task (CTRLsum). 
For both PEGASUS and BART, we use two different kinds of input, source document only (-Doc) and with the controlling aspects as the prefix (-CA+Doc).

Besides, we also compare our method with the three backbone models, when they are extended with a widely used plug-and-play module (PPLM, ~\citet{pplm}. The details of the models can be found in Appendix~\ref{sec:bsl_models}.

\section{Experiments and Analysis}

\begin{table*}[h!]
    \centering
    \small
    \begin{tabular}{l|cccc|cccc}
    \toprule
    \multirow{2}{*}{Model}     
    &\multicolumn{4}{c|}{
    NEWTS}&\multicolumn{4}{c}{
    EntSUM}  \\
    & R-1&R-2&R-LSum&BERTScore&R-1&R-2&R-LSum&BERTScore\\
    \midrule
    General Summ&28.82&7.62&23.55&84.99&27.51&9.79&18.26&84.07\\
    \midrule
    PEGASUS-Doc &33.21&11.41&29.06&\textbf{85.55}
                &36.44&18.75&32.76&85.72\\
    PEGASUS-CA+Doc&\textbf{33.58}&\textbf{11.45}&29.02&85.20
                    &36.99&19.08&33.19&85.69\\
    PEGASUS+PPLM(Val)&32.11&10.88&28.08&85.32&34.90&17.10&31.17&85.29\\
    PEGASUS+RelAttn(Val)&\textbf{33.37}&\textbf{11.50}&\textbf{29.28}&\textbf{85.59}
            &\textbf{37.49}$\dagger$&\textbf{19.84}$\dagger$&\textbf{33.84}$\dagger$&\textbf{85.92}\\
    PEGASUS+RelAttn(OS)&33.23&11.33&\textbf{29.20}&85.43
            &\textbf{39.02}$\dagger$&\textbf{21.43}$\dagger$&\textbf{35.36}$\dagger$&\textbf{86.14}\\
    \textit{PEGASUS+RelAttn(OS-oracle)}&36.87&13.53&32.35&85.97&44.90&27.33&41.14&87.10\\
    \midrule
BART-Doc  &  33.09&10.67&28.98&\textbf{86.01}
        &30.50&12.10&27.33&85.16\\
BART-CA+Doc  & 32.90&11.01&29.48&85.98
        &30.73&12.46&27.59&85.17 \\
    BART+PPLM(Val)&31.61&10.30&28.67&85.70&29.22&10.56&26.25&84.67\\
    BART+RelAttn (Val)&\textbf{33.10}&\textbf{11.11}&\textbf{29.70}&\textbf{86.01}
            &\textbf{33.47}$\dagger$&\textbf{16.07}$\dagger$&\textbf{30.55}$\dagger$&\textbf{85.58}\\
    BART+RelAttn(OS)&\textbf{33.34}$\dagger$&\textbf{11.19}&\textbf{30.09}$\dagger$&\textbf{86.06}
            &\textbf{34.42}$\dagger$&\textbf{16.23}$\dagger$&\textbf{31.21}$\dagger$&\textbf{85.88}\\
    \textit{BART+RelAttn(OS-oracle)}&39.48&14.95&35.58&86.81&44.83&27.76&41.57&87.54\\
    \midrule
    CTRLsum* &\textbf{32.57}&\textbf{9.58}&\textbf{29.06}&\textbf{85.52}
            & 35.97&\textbf{18.56}&32.69&85.96\\
    CTRLsum+PPLM(Val)&30.32&8.31&27.23&85.26&32.40&14.33&29.18&85.10\\
    CTRLsum+RelAttn(Val)&32.12&9.17&28.45&\textbf{85.51}
            &\textbf{36.14}&18.40&\textbf{32.82}&\textbf{86.03}\\
    CTRLsum+RelAttn(OS)&\textbf{32.38}&\textbf{9.53}&\textbf{28.91}&85.40
            &\textbf{37.24}$\dagger$&\textbf{19.27}$\dagger$&\textbf{33.71}$\dagger$&\textbf{86.20}\\
    \textit{CTRLsum+RelAttn(OS-oracle)}&37.54&12.01&33.17&85.92&45.25&27.05&41.34&87.47\\
    \bottomrule
    \end{tabular}
    \caption{Results on the three scenarios in zero-shot settings with three different backbone models. The top-2 performers (except the oracle ones) are bold. $\dagger$ indicates that the model is significantly better than the baseline models with $p<0.05$ by bootstrap significant test~\cite{graham-etal-2014-randomized}. }
    \label{tab:zero-shot}
\end{table*}
\begin{table}[]
    \centering
    \small
    \begin{tabular}{l|c|c}
     \toprule
     Model    &  NEWTS & EntSUM\\
     \midrule
    PEGASUS-Doc&26.91&21.39\\
    PEGASUS-CA+Doc&\textbf{32.20}&24.41\\
    PEGASUS+RelAttn &28.79&\textbf{25.78}\\
    \midrule
    BART-Doc&26.29&20.42\\
    BART-CA+Doc&26.60&20.70\\
    BART+RelAttn &\textbf{28.90}&\textbf{25.95}\\
    \midrule
    CTRLsum&42.02&30.52\\
    CTRLsum+RelAttn &\textbf{43.56}&\textbf{35.72}\\
    \bottomrule
    \end{tabular}
    \caption{SimCSE between the generated summary and the controlling aspects.}
    \label{tab:simcse_ca}
\end{table}

\subsection{Settings and Implementation}
\label{sec:settings_and_implementation}
For all the pre-trained models (PEGASUS, BART-CNNDM, and CTRLsum) used in this paper, we directly use the publicly available checkpoints\footnote{\url{https://huggingface.co/models}}. 

In the zero-shot setting, the relevance weight $w_{rel}$ is selected 
in three ways, as listed below,
from a candidate set, which contains all the 30 numbers within the range $[0.01,0.30]$ with a step size $0.01$. And we also add several heuristic constraints to ensure the summaries generated by $w_{rel}$ in the candidate sets are reasonable: we remove all the candidate summaries with too many repeated words and the summaries distracted too much \footnote{By practice, we remove the summaries with unique word ratios less than $0.6$ and with word overlap less than $0.2$ with the generated summary without control.}, and we also remove all the duplicated summaries. \footnote{We set the upper bound to be $0.30$ for computational efficiency, and by observation, for most of the data examples, the real upper bound for the filtered candidate set is below $0.30$.}

 \textbf{RelAttn(Val)}: $w_{rel}$ is a fixed number shared for all the data examples in the dataset, and it is set based on the performances on a randomly sampled validation set with size 100. 
 
\textbf{RelAttn(OS)}: $w_{rel}$ is determined using the online selection algorithm introduced in Section \ref{sec:os_algorithm}, and it can be different for each single data example. We use ROUGE-1 as the alignment score in the OS algorithm. 
    
\textbf{RelAttn(OS-oracle)}: $w_{rel}$ is set to be the number %
with which the summaries have the best ROUGE score with the ground-truth summaries, and it can be different for each single data example. As such, it shows the \textit{empirical} upper bound of the RelAttn model in the zero-shot setting.\footnote{
    Since we only compute it for 30 values, an even higher bound could be possible.
    }

More detailed settings can be found in Appendix~\ref{sec:settings_appendix}.
\subsection{Zero-Shot Results}
We first evaluate all the models on the two datasets in the zero-shot settings with the ROUGE scores and BERTScore between the generated summaries and the ground-truth summaries. The results are shown in Table~\ref{tab:zero-shot}. The top blocks show the scores between the controlled and general summaries, and the following blocks show the performances of three backbone models, PEGASUS, BART, and CTRLsum. 

The scores between the general and controlled summaries are quite low for both datasets, indicating that the controlled summaries are mostly different from the general summaries.

\textbf{RelAttn generally helps.} With the proposed RelAttn module (-Doc v.s. +RelAttn), the performances of all three models on both datasets improve except for the CTRLsum on the NEWTS dataset, and the improvements are mostly significant with $p<0.05$. In essence, the RelAttn module is shown to be more effective for all three models on the EntSUM dataset, and the reason might be that as the dataset is more extractive, paying attention to the correct context brings a more obvious improvement.

\textbf{Simply adding the controlling aspect as a prefix provides limited help.} Comparing the models (PEGASUS and BART) with different inputs, adding the controlling aspects as prefix (-CA+DOC v.s. -DOC) brings improvements on the EntSUM dataset for both models, and on the NEWTS for BART regarding ROUGE scores. However, the improvements on the ROUGE scores are limited, and the BERTScore even drops (within a small margin) in most of the cases.

\textbf{Pre-training for guided summarization does not lead to a performance gain on the controllable summarization task.}
As a pre-trained guided-summarizer, CTLRSum does not perform as well as the other two models on the NEWTS dataset, and also has a lower score compared with PEGASUS on the EntSUM dataset, which shows that the model pre-trained for the guided summarization task may not be directly applied to the controllable summarization task, in the case where the relevant content and generally salient content are different.

\textbf{Flexible $w_{rel}$ by unsupervised OS algorithm works better than a fixed number set based on a validation set.} In five out of six sets of experiments (except for the PEGASUS model on the NEWTS dataset), picking $w_{rel}$ by the unsupervised OS algorithm for each data example has a higher performance than using a fixed number based on a small validation set. This suggests that different data example requires a different degree of controlling and changing, which could be effectively determined by the online selection algorithm.

\textbf{The performances of all the models still have a large room to improve with improved selection algorithm for  $w_{rel}$.} For all the models on both datasets, the results of RelAttn(OS) are still much less than the results of RelAttn(OS-Oracle), which are the empirical upper bounds for the RelAttn module in the zero-shot settings. 

\textbf{The generated summaries have higher similarity with the given controlling aspects when RelAttn is used.} To evaluate how relevant the summaries are with the controlling aspects, we compute the SimCSE score between the generated summaries and the controlling aspects, and the results are shown in Table~\ref{tab:simcse_ca}. It can be seen that the RelAttn module can effectively improve the similarity between the generated summaries and the controlling aspects in most of the cases, except when applied to the PEGASUS model on the NEWTS dataset. Plausibly, a possible explanation might be that the pre-trained PEGASUS model tends to directly copy the sentences from the source documents to the summary, and based on the way the pre-trained data is built, the first sentences are frequently copied. As a result, the prefix (controlling aspects) is also copied to the summary, which will therefore have high similarity with the controlling aspects. 

\subsection{Analysis on the Degree of Control}
\begin{figure}[h!]
    \centering
    \includegraphics[width=\linewidth]{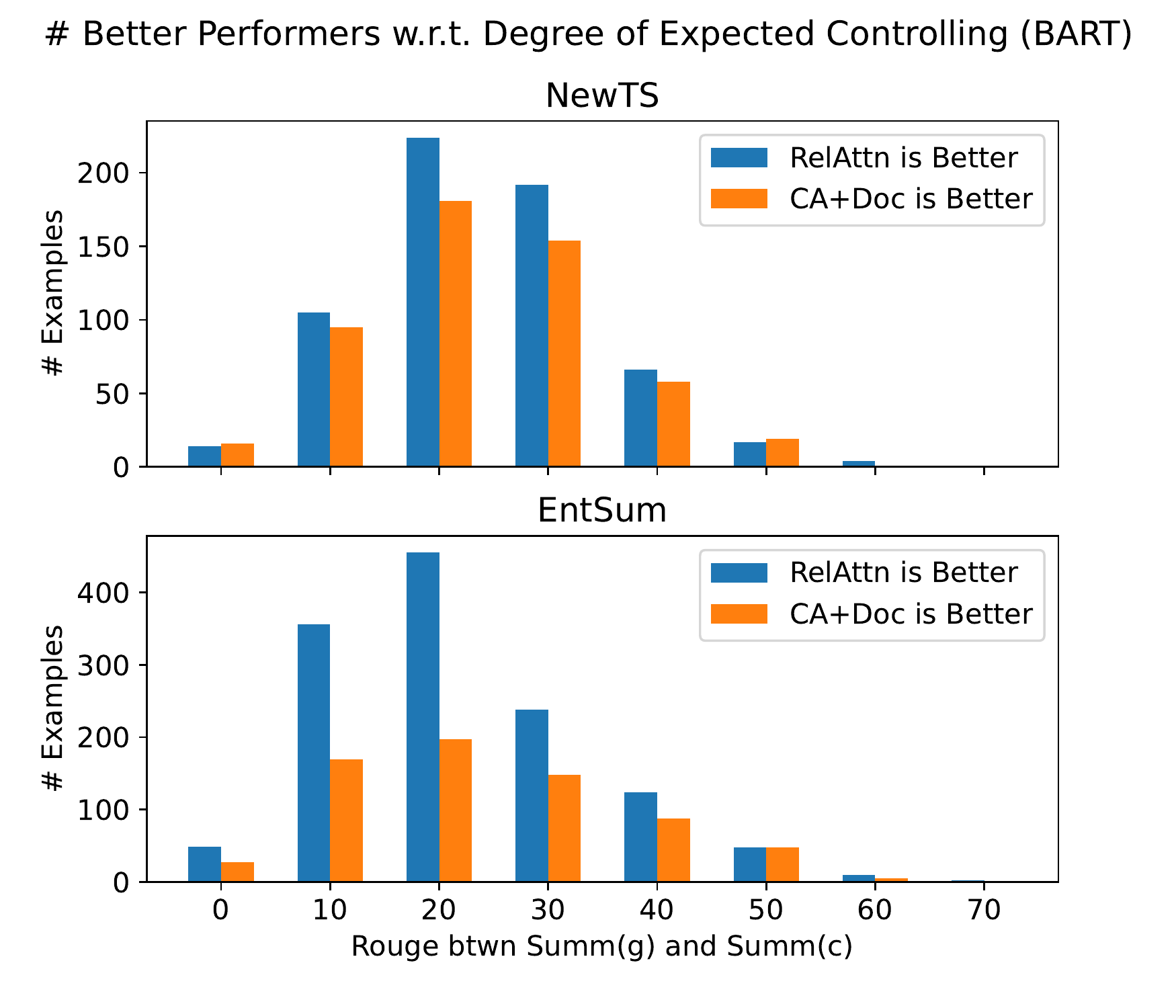}
    \caption{The number of better performers among CA+Doc and RelAttn(OS) regarding different degrees of expected control (measured by the ROUGE-1 score between the general ground-truth summaries and the controlled ground-truth summaries) on both datasets (with BART as the backbone model).}
    \label{fig:better_performer_bart}
\end{figure}

\begin{figure}[h!]
    \centering
    \includegraphics[width=\linewidth]{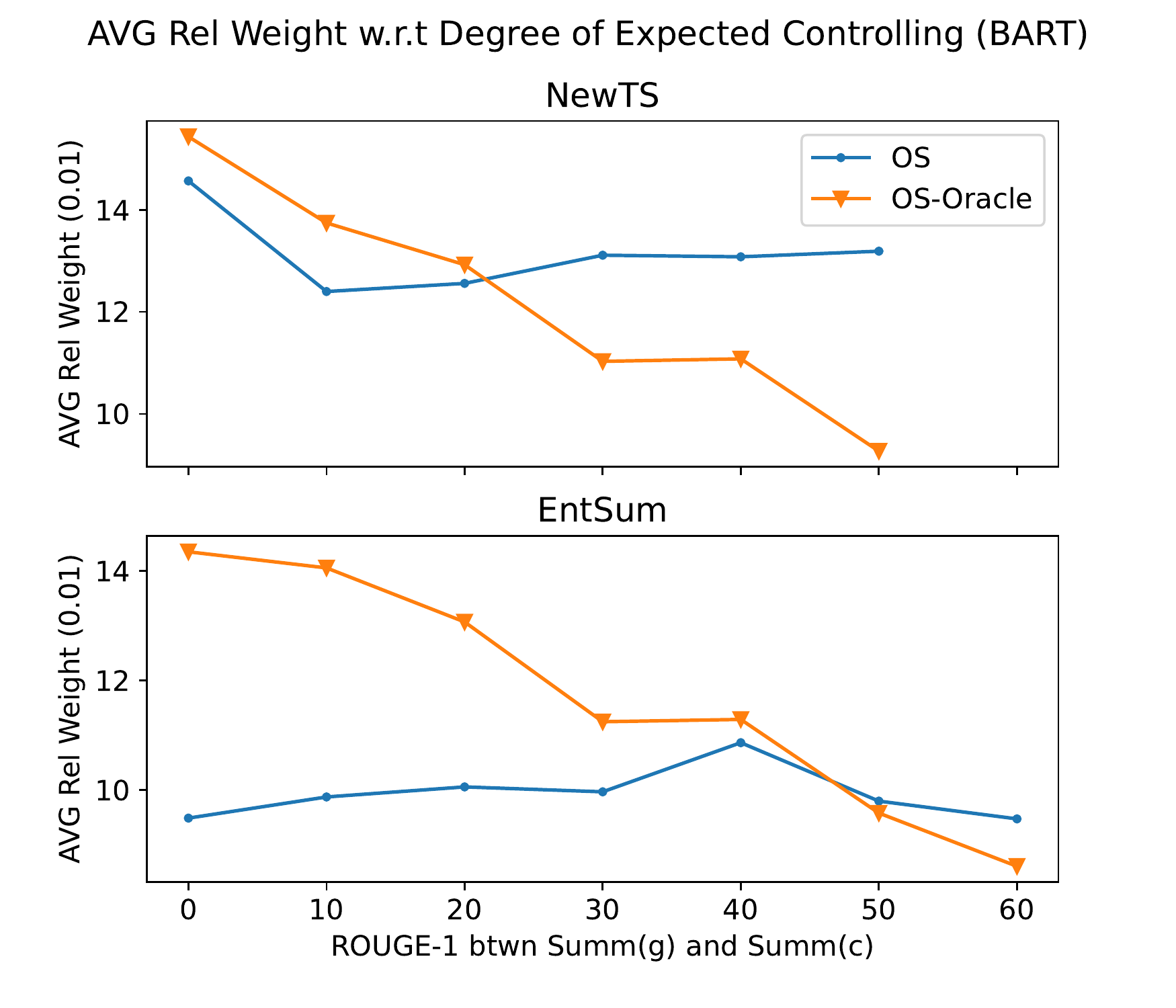}
    \caption{The avg $w_{rel}$ of  OS and -Oracle regarding the different degrees of expected control on both datasets (with BART as the backbone model).}
    \label{fig:w_rel_comparison_bart}
\end{figure}
We further explore the effectiveness of our method regarding different levels of control needed. We use the similarity (ROUGE) between gold-standard controlled summaries and general summaries to measure the degree of expected control, and intuitively, a higher degree of control is needed if the controlled summaries are less similar to the general summaries. 
Specifically, we split all the data examples into bins according to the ROUGE-1 scores, and analyze the performances of our method on the data examples in each bin.

We show the number of better performers among RelAttn (OS) and prefix-based method -CA+DOC regarding different degrees of controlling in Figure~\ref{fig:better_performer_bart}. Based on the results, RelAttn tends to perform better when more controlling is needed for both datasets, especially when the ROUGE-1 between the general summaries and controlled summaries are within $[10,40]$. 

We then explore the value of $w_{rel}$ regarding the different degrees of control in Figure~\ref{fig:w_rel_comparison_bart}. By design, $w_{rel}$ is supposed to be higher for the data examples requiring more control. The oracle $w_{rel}$ (by OS-Oracle) follows the expected trend on both datasets.\footnote{The results of the bins with less than 10 data examples are omitted.} However, the average values for $w_{rel}$ selected by the OS algorithm are similar across all the bins, which are around $12-14$ on NEWTS and around $10$ for EntSUM. A possible reason is that we select the proper $w_{rel}$ with the `central' summary among all the candidates for each example, which is likely to be the `central value' in the candidate set. Improving the unsupervised algorithm by taking into consideration the distance between salient and relevant content is left as future work.
 
\begin{table*}[t!]
    \centering
    \small
    \begin{tabular}{l|ccc|ccc}
    \toprule
    \multirow{3}{*}{Model}    
    &\multicolumn{3}{c|}{
    NEWTS}&\multicolumn{3}{c}{
    EntSUM}  \\
    \cline{2-7}
    & R-1&R-2&R-L&R-1&R-2&R-L\\
    \midrule
    Pgeasus-CA+Doc&\textbf{33.63}$\dagger$&\textbf{11.42}&\textbf{29.06}&37.12&19.31&33.30\\
    PEGASUS+RelAttn (Ours)&33.19&11.32&29.03&\textbf{37.66}$\dagger$&\textbf{20.25}$\dagger$&\textbf{34.03}$\dagger$\\
    \midrule
BART-CA+Doc  & 33.65&11.26&30.09&39.45&23.32&36.23 \\
BART+RelAttn (Ours)&\textbf{34.36}$\dagger$&\textbf{11.81}$\dagger$&\textbf{30.67}$\dagger$&\textbf{40.27}$\dagger$&\textbf{24.63}$\dagger$&\textbf{37.03}$\dagger$ \\
    \midrule
CTRLsum&33.54& 11.01&29.92&51.57& 39.83& 49.00\\
    CTRLsum+RelAttn (Ours)&\textbf{33.88}$\dagger$&\textbf{11.07}&\textbf{30.18}$\dagger$&\textbf{52.42}$\dagger$&\textbf{40.65}$\dagger$&\textbf{49.82}$\dagger$\\
    \bottomrule
    \end{tabular}
    \caption{Results on the two datasets in few-shot settings with three different backbone models. We train all the models on 10 data examples for 5 times, with the same set of random seeds, and the results shown in the table are the mean ROUGE scores. $\dagger$ represents the score is significantly better than the other score in the same block with $p<0.05$ by bootstrap significant test\cite{graham-etal-2014-randomized}.}
    \label{tab:few-shot}
\end{table*}
\begin{table*}[h!]
    \centering
    \small
    \begin{tabular}{p{0.2\linewidth}|p{0.8\linewidth}}
    \toprule
  Model Type &Summary\\
  \midrule
      Ground-truth& The reporter, \textcolor{red}{ James Buckley}, 50, of Brookhaven, who has worked for more than 20 years at WALK-FM (97.5), was arraigned on Tuesday morning in United States District Court in Central Islip. Allen Bode, an assistant United States attorney, said in a telephone interview that \textcolor{red}{Mr. Buckley} was caught trading pornography, including some hard-core material involving toddlers, with undercover agents in Germany and the United States. \textcolor{red}{Mr. Buckley} faces up to 19 years in federal prison if convicted.\\
  \midrule
  BART-Doc \newline ($w_{rel}=0$)&24 men accused of soliciting sex from detectives posing as children in computer chat rooms.
In a separate case, a traffic and weather reporter for a Long Island radio station was indicted on federal charges of possessing child pornography.
All were charged with attempting to disseminate indecent material to a minor, a felony with a maximum penalty of four years in prison.
The 24 cases together constitute the largest child sex crackdown ever in Suffolk, the police said.\\
\midrule
    BART+RelAttn (OS-Oracle)\newline ($w_{rel}=0.12$) &\textcolor{red}{Buckley} faces up to 19 years in federal prison if convicted.\textcolor{red}{Buckley} has worked for more than 20 years at WALK-FM (97.5) \textcolor{red}{Buckley's} court-appointed defender, Randi Chavis, said in a telephone interview that he pleaded not guilty. \textcolor{red}{Buckley} was caught trading pornography, including some hard-core material involving toddlers, with undercover agents in Germany and the United States.\textcolor{red}{Buckley's} attorney declined to comment.\\
\midrule
BART+RelAttn (OS)\newline ($w_{rel}=0.15$)&\textcolor{red}{Buckley} faces up to \textit{...(same as above)} declined to comment.\textcolor{red}{Buckley} is one of 11 men arrested after sending sexually explicit messages.\\
\midrule
BART+RelAttn\newline ($w_{rel}=0.30$)& \textcolor{red}{Buckley Buckley, Buckley Buckley Buckley} is \textcolor{red}{Buckley Buckley's Buckley Buckley}.\textcolor{red}{Buckley Buckley} faces \textcolor{red}{Buckley Buckley} was ...\\
\midrule
CTRLsum&\textcolor{red}{James Buckley}, 50, of Brookhaven, faces up to 19 years in federal prison if convicted.
Police say they created online profiles and screen names for fictitious boys and girls.
Police commissioner: ''I consider this the sleaziest kind of investigation, but I consider it necessary'' Police say 11 others were arrested after sending sexually explicit messages or photographs online; their identities were provided by the Internet service providers under subpoena.
autoimmune autoimmune autoimmune disease.
autoimmune disease, autoimmune disease of the mind, immune system, and body.\\
\midrule 
CTRLsum+RelAttn(OS)\newline ($w_{rel}=0.05$)&\textcolor{red}{James Buckley}, 50, of Brookhaven, faces up to 19 years in federal prison if convicted.
\textcolor{red}{Buckley} was caught trading pornography, including hard-core material involving toddlers, with undercover agents in Germany and the U.S. Police say they created online profiles and screen names for fictitious boys and girls, ages 12 to 14, and visited chat rooms operated by AOL, Yahoo and other Internet service providers.
Police commissioner: ''I consider this the sleaziest kind of investigation, but I consider it necessary''\\
    \bottomrule
    \end{tabular}
    \caption{An example from the EntSUM dataset  (\textit{doc\_id: 1739833}) with the entity \textbf{James Buckley}, and the full version can be found in Table~\ref{tab:example_full} in the Appendix. }
    \label{tab:example}
\end{table*}
\subsection{Few Shot}
We train all three models on both datasets with 10 data examples for 5 times with the same set of random seeds and report the averaged ROUGE scores in Table~\ref{tab:few-shot}. 

On five out of six sets of experiments, our method shows significant improvements over the prefix-based method with only 10 training examples. Comparing the three backbone models, CTRLsum makes the largest improvements, and PEGASUS makes the least improvements. The reason might be that the guided summarizer is trained in the way that is closest to the controllable summarization task, while the pre-trained summarization model, which has not been trained on any labeled data, needs more in-domain data to train. 

\subsection{Qualitative Analysis}
We show the results of a real example for zero-shot settings from the EntSUM dataset in Table~\ref{tab:example}. The news article is about a set of crimes conducted in a computer chat room, and the given entity is one of the accused suspects, called James Buckley. The summary generated by the general summarizer (BART-Doc) does not contain the given entity, and it is generally about the set of crimes, without focusing on any particular one. With the RelAttn module, the model takes consideration of the given entity, thus the generated summaries focus on the content related to the given entity (BART+RelAttn). However, when $w_{rel}$ is too large, the generated summary only repeats the given entity rather than a fluent text, which shows the necessity of selecting a proper $w_{rel}$. The trained guided-summarization model (CTRLsum) can generate a summary related to the given entity, however, it simply generates one related sentence, with the remaining summaries still focusing on the general salient content other than the relevant content. With the RelAttn module, the generated summaries focus more on the given entity.

\section{Conclusion and Future Work}
In this work, we propose a plug-and-play module, RelAttn, to adapt %
general summarization models to the content-controllable summarization task in zero/few-shot settings. In essence, RelAttn helps the model focus on the content more relevant to the given controlling aspects by steering the attention weights in the cross attention. Remarkably, it can be injected into any transformer-based sequence-to-sequence model. %
An online adaptive parameter selection algorithm is used in the zero-shot setting to estimate the degree of control, and the module is also trainable in the few-shot settings. We conduct experiments on two datasets with three summarization models as backbone models. The results show that RelAttn effectively adapts the general summarizers to the content-controllable summarization task in both zero- and few-shot settings. Tellingly, further analysis suggests that the module is more helpful when more control is needed in the zero-shot settings.

Despite the performance gain over the baseline models, there is still a large room for improvement with a better way to estimate the degree of control in the zero-shot setting, thus one of the potential future works is to improve the unsupervised algorithm by taking considerations of the distance between the salient content and relevant content. Furthermore, due to the nature of the RelAttn module, it also supports \textit{abstractive} content-controlling. Instead of using the similarities between the tokens of controlling aspects and the source documents, one could also build the Relevance Attention using the similarity between a controlling embedding (e.g. different aspects for reviews or different topic representations) and the source documents, which is left as future work.
\section*{Acknowledgement}
We thank Song Wang and Jie Mei for valuable discussions.

\bibliography{anthology,custom}
\bibliographystyle{acl_natbib}

\appendix
\section{Details of the Datasets}
\label{sec:dataset_details_appendix}
We show two examples from the two datasets in Table~\ref{tab:dataset_example}
\begin{table*}[]
    \centering
    \small
    \begin{tabular}{p{\linewidth}}
 
    \toprule
    NEWTS\\
    \hline
    \textcolor{blue}{Source}: An American tourist has spent the night stranded in the Blue Mountains, west of Sydney, after she fell 15 metres off a cliff while bushwalking
    \newline ...\newline 
    'But due to the terrain in the prevailing weather that plan was aborted.' Rescue teams had to wait for the fog to lift so they could winch the woman out via a helicopter.
    \tabularnewline
    \textcolor{OliveGreen}{Summary}: Foggy weather conditions made it difficult to rescue a stranded hiker. Helicopters cannot fly with such low visibility. The weather also blocked out sunlight.
    \tabularnewline
    \textcolor{red}{Controlling Aspects}: snow, weather, cold, winter, temperatures, conditions, hot, morning, expected, parts\\
    \midrule
    EntSUM\\
    \hline
    \textcolor{blue}{Source}: A lieutenant colonel in the Army Reserve, a firefighter and a college student are among 24 Manhattan and Long Island men accused of soliciting sex from detectives posing as children in computer chat rooms during a monthlong sting operation.
    \newline ... \newline
    The reporter, James Buckley, 50, of Brookhaven, who has worked for more than 20 years at WALK-FM (97.5), was arraigned on Tuesday morning in United States District Court in Central Islip. Magistrate Judge Arlene R. Lindsay ordered him held under house arrest on \$350,000 bond. 
    \newline...\newline
Commissioner Dormer said the detectives stayed in character, as naïve children.
''The predators seduced the youngsters; the youngsters did not seduce the predators,'' he said.
    \tabularnewline
    \textcolor{OliveGreen}{Summary}: The reporter, James Buckley, 50, of Brookhaven, who has worked for more than 20 years at WALK-FM (97.5), was arraigned on Tuesday morning in United States District Court in Central Islip.
    ...
    Mr. Buckley faces up to 19 years in federal prison if convicted.
    \tabularnewline
    \textcolor{red}{Controlling Aspects}: James Buckley\\
    \bottomrule
    \end{tabular}
    \caption{Examples from both datasets.}
    \label{tab:dataset_example}
\end{table*}
\section{Details of the Baseline Models}
\label{sec:bsl_models}
To evaluate the effectiveness of our method, we use three pre-trained models as the backbone model, including a pre-trained model for general summarization (PEGASUS), a model fine-tuned for the general summarization task (BART-CNNDM), and a model trained for conditional summarization task (CTRLsum). As both PEGASUS and BART-CNNDM are (pre-)trained for the general summarization task, we prepend the source documents with the given controlling aspects as prefixes to fit the model to the controlled summarization task, followed by a special token and the source document as the inputs to the models. Besides, we also compare our method with the three backbone models, when they are extended with a widely used plug-and-play module (PPLM)~\cite{pplm}. 
\paragraph{PEGASUS~\cite{pegasus}} is a pre-trained generative model tailored for abstractive summarization, with the objective of generating the mask-out salient sentences in the source documents. It shows a good performance for the general summarization task in zero-shot and few-shot settings.
\paragraph{BART-CNNDM~\cite{bart}} is a pre-trained generative language model, fine-tuned on the CNNDM dataset~\cite{cnndm} for the general summarization task, which achieves the state-of-the-art on the CNNDM dataset.
\paragraph{CTRLsum~\cite{ctrlsum}} is a pre-trained model for guided summarization, which is trained with the oracle keyphrases on the CNNDM dataset~\cite{cnndm}. Specifically, in the training stage, the oracle keyphrases (the matched keyphrases between the source document and the ground truth summary) is prepended to the source documents as the input, and in the inference stage, the keywords can either be automatically generated or human given.

\paragraph{PPLM~\cite{pplm}} is a plug-and-play language model for conditional generation, which can be applied to any transformer-based generative model. It combines the pre-trained generative language model with one or more trained simple attribute models that guide text generation without any further training of the LM. Specifically, at each generation step, it updates the language model's hidden states using the attribute model's gradients with the current generated text as the input. The updates toward the direction to the combination of  higher log-likelihood (LL) of the attribute $a$ under the conditional attribute model $P(a|x)$ and higher LL of the unmodified language model $P(x)$. In this work, as we focus on content-controlling, we simply use the bag-of-words of the controlling factor as the attribute model.
We apply the plug-and-play component to all the aforementioned models. As the PPLM module can not be trained, we only evaluate them in the zero-shot setting.
\section{Settings and Implementations}
\label{sec:settings_appendix}
For the PPLM model, we do a hyper-parameter search on a randomly sampled validation set with size 100 for all the backbone models on both datasets. The hyper-parameters are $\gamma_{gm}$ and step size, which are searched within $\{0.65,0.75,0.85,0.95\}$ and $\{e^{-2},e^{-3},e^{-4},e^{-5}\}$, respectvely.

For all the models, we use the default settings for the generation step as in the original model, and the length limit of the decoder is set to $56/142$, $86/172$ for NEWTS and EntSUM respectively based on the average length of the two datasets.
\section{Qualitative Analysis}
In Table~\ref{tab:example_full}, we show the full example as shown in Table~\ref{tab:example}. It can be found that the PEGASUS model directly copies the first sentence (along with the prefix) to the summary, and with RelAttn, the model additionally generates a relevant sentence.
\begin{table*}[]
    \centering
    \small
    \begin{tabular}{p{0.2\linewidth}|p{0.8\linewidth}}
    \toprule
  Model Type &Summary\\
  \midrule
      gt& The reporter, \textcolor{red}{ James Buckley}, 50, of Brookhaven, who has worked for more than 20 years at WALK-FM (97.5), was arraigned on Tuesday morning in United States District Court in Central Islip. Allen Bode, an assistant United States attorney, said in a telephone interview that \textcolor{red}{Mr. Buckley} was caught trading pornography, including some hard-core material involving toddlers, with undercover agents in Germany and the United States. \textcolor{red}{Mr. Buckley} faces up to 19 years in federal prison if convicted.\\
  \midrule
  PEGASUS-DOC&A lieutenant colonel in the Army Reserve, a firefighter and a college student are among 24 Manhattan and Long Island men accused of soliciting sex from detectives posing as children in computer chat rooms during a monthlong sting operation, the Suffolk County police said on Tuesday.
''Let this be a wake-up call to parents to monitor our children's chat rooms,'' said County Executive Steve Levy at a news conference at police headquarters in Yaphank.
The Suffolk police commissioner, Richard Dormer, who was also at the news conference, said, ''I consider this the sleaziest kind of investigation, but I consider it necessary.''
In a separate case, a traffic and weather reporter for a Long Island radio station was indicted on federal charges of possessing child pornography.
Detectives in the Police Department's computer crimes unit who were at the\\
\midrule
  PEGASUS-CA+DOC&\textcolor{red}{James Buckley} => A lieutenant colonel in the Army Reserve, a firefighter and a college student are among 24 Manhattan and Long Island men ... (same as above)\\
\midrule
PEGASUS+RelAttn(OS)\newline ($w_{rel}=0.10$) &A lieutenant colonel in the Army Reserve, a 
...(same as above)
federal charges of possessing child pornography.
\textcolor{red}{Buckley Buckley Buckley} was caught trading pornographic messages, including some hard-core\\
\midrule
  BART-Doc&24 men accused of soliciting sex from detectives posing as children in computer chat rooms.
In separate case, a traffic and weather reporter for a Long Island radio station indicted on federal charges of possessing child pornography.
All were charged with attempting to disseminate indecent material to a minor, a felony with a maximum penalty of four years in prison.
The 24 cases together constitute the largest child sex crackdown ever in Suffolk, the police said.\\
\midrule
BART-CA+DOC&24 men accused of soliciting sex from detectives posing as children in computer chat rooms.
Police commissioner: ''I consider this the sleaziest kind of investigation, but I consider it necessary'' In separate case, Long Island radio reporter indicted on federal charges of possessing child pornography.\\
\midrule
    BART+RelAttn (OS-Best)\newline ($w_{rel}=0.12$) &\textcolor{red}{Buckley} faces up to 19 years in federal prison if convicted.Buckley has worked for more than 20 years at WALK-FM (97.5) \textcolor{red}{Buckley's} court-appointed defender, Randi Chavis, said in a telephone interview that he pleaded not guilty. \textcolor{red}{Buckley} was caught trading pornography, including some hard-core material involving toddlers, with undercover agents in Germany and the United States.\textcolor{red}{Buckley's} attorney declined to comment.\\
\midrule
BART+RelAttn (OS)\newline ($w_{rel}=0.15$)&\textcolor{red}{Buckley} faces up to \textit{...(same as above)} declined to comment.\textcolor{red}{Buckley} is one of 11 men arrested after sending sexually explicit messages.\\
\midrule
    BART+RelAttn (Val)\newline ($w_{rel}=0.17$)& \textcolor{red}{Buckley} faces up to \textit{...(same as above)} undercover agents in Germany and the United States.\textcolor{red}{Buckley's} attorney, \textcolor{red}{Buckley Buckley}, declined to comment.\textcolor{red}{Buckley} is one of 11 men arrested after sending sexually explicit messages.\\
\midrule
BART+RelAttn\newline ($w_{rel}=0.30$)& \textcolor{red}{Buckley Buckley, Buckley Buckley Buckley} is \textcolor{red}{Buckley Buckley's Buckley Buckley}.\textcolor{red}{Buckley Buckley} faces \textcolor{red}{Buckley Buckley} was ...\\
\midrule
CTRLsum&\textcolor{red}{James Buckley}, 50, of Brookhaven, faces up to 19 years in federal prison if convicted.
Police say they created online profiles and screen names for fictitious boys and girls.
Police commissioner: ''I consider this the sleaziest kind of investigation, but I consider it necessary'' Police say 11 others were arrested after sending sexually explicit messages or photographs online; their identities were provided by the Internet service providers under subpoena.
autoimmune autoimmune autoimmune disease.
autoimmune disease, autoimmune disease of the mind, immune system, and body.\\
\midrule 
CTRLsum+RelAttn(OS)\newline ($w_{rel}=0.05$)&\textcolor{red}{James Buckley}, 50, of Brookhaven, faces up to 19 years in federal prison if convicted.
\textcolor{red}{Buckley} was caught trading pornography, including hard-core material involving toddlers, with undercover agents in Germany and the U.S. Police say they created online profiles and screen names for fictitious boys and girls, ages 12 to 14, and visited chat rooms operated by AOL, Yahoo and other Internet service providers.
Police commissioner: ''I consider this the sleaziest kind of investigation, but I consider it necessary''\\
    \bottomrule
    \end{tabular}
    \caption{The full example from the EntSUM dataset (\textit{doc\_id: 1739833}) with the entity \textbf{James Buckley}. }
    \label{tab:example_full}
\end{table*}
\end{document}